# Robots Learn Social Skills: End-to-End Learning of Co-Speech Gesture Generation for Humanoid Robots

Youngwoo Yoon, Woo-Ri Ko, Minsu Jang, Jaeyeon Lee, Jaehong Kim, and Geehyuk Lee

*Abstract*— Co-speech gestures enhance interaction experiences between humans as well as between humans and robots. Existing robots use rule-based speech-gesture association, but this requires human labor and prior knowledge of experts to be implemented. We present a learning-based co-speech gesture generation that is learned from 52 h of TED talks. The proposed end-to-end neural network model consists of an encoder for speech text understanding and a decoder to generate a sequence of gestures. The model successfully produces various gestures including iconic, metaphoric, deictic, and beat gestures. In a subjective evaluation, participants reported that the gestures were human-like and matched the speech content. We also demonstrate a co-speech gesture with a NAO robot working in real time.

## I. INTRODUCTION

The social intelligence of artificial agents is getting attention as social robots rise and people interact more with robots. Evaluation of Social Interaction (ESI), which is a standardized assessment tool for humans, present key social skills including approaches, speaking, turn-taking, gaze, and gesticulation [1]. In the present paper, we focus on gesticulation, particularly co-speech gestures for robots. People use co-speech gestures when they talk to others to emphasize speech, show intention, or describe something vividly [2]. Many social science studies proved the positive effects of co-speech gestures [3], [4], and a neuroscience study also indicates that co-speech gestures help discourse comprehension [5]. Recent social robots such as Pepper [6] and RoboThespian [7] are able to make human-like co-speech gestures, but the gestures are crafted by human experts. While manually crafted gestures are natural and human-like, there is a limitation in that only gestures considered in the design stage can be performed. Furthermore, building associations between gestures and speech words requires significant human labor.

This paper presents a learning-based co-speech gesture generation method for humanoid robots (Fig. 1). Robots can learn co-speech gestures from human behaviors as we ourselves do. Mimicking human gestures is a viable strategy for humanoid robots since they have similar appearances and control joints. We propose an end-to-end model that produces co-speech gestures, specifically temporal sequences of upper-body poses, for given speech in natural language. The model

This work was supported by the ICT R&D program of MSIP/IITP. [2017-0-00162, Development of Human-care Robot Technology for Aging Society].

Y. Yoon is with HCI Lab, School of Computing, KAIST and Intelligent Cognitive Technology Research Department, ETRI, Republic of Korea (youngwoo@etri.re.kr).

W. Ko, M. Jang, J. Lee and J. Kim are with Intelligent Cognitive Technology Research Department, ETRI, Republic of Korea ({wrko, minsu, leejy, jhkim504}@etri.re.kr).

G. Lee is with HCI Lab, School of Computing, KAIST, Republic of Korea (geehyuk@gmail.com).

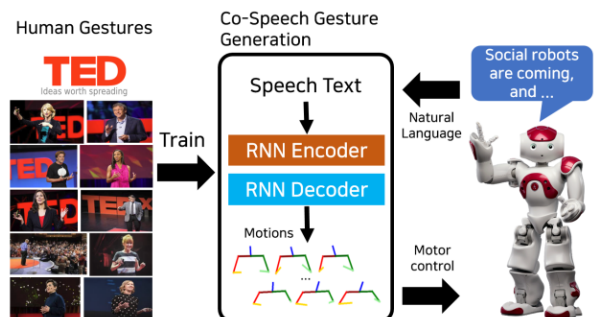

Figure 1. We address a problem of making co-speech gestures for a given speech text. The proposed model generates a sequence of upper-body poses, and it is trained from human gestures in TED talks.

uses a sequence-to-sequence architecture consisting of an encoder and decoder, mapping speech words to gestures. Any prior knowledge about gesticulation and speech-gesture mapping is not imposed. The model is trained only from human demonstrations, and we expect the model to make proper and various co-speech motions of iconic, metaphoric, deictic, and beat gestures [2].

Our contribution is three fold. First, we present a new large-scale dataset for co-speech gesture studies. It contains 52 h of videos of human gestures, and it comes with speech transcripts. The dataset is collected from TED talks in which various people make speeches on various topics, so it is beneficial not only to the HRI community but also to the social science community. Second, a novel gesture generation model designed for end-to-end learning is proposed. The model inputs natural language and outputs frame-by-frame poses. Without prior knowledge, the model generates several types of gestures freely for speech text never seen before. Third, we bring generated gesture motions into reality by implementing a robot prototype that makes gestures while speaking in real time. The conversion of 2D poses to 3D poses and aligning speech audio and motions are investigated.

The next section gives related works compared with our approach. Section III and IV describe the collected TED gesture dataset and the proposed gesture generation method, respectively. Subjective evaluation results follow in Section V. We describe details of the robot prototype in Section VI. Section VII presents discussions and limitations.

## II. RELATED WORKS

### A. Automatic Co-Speech Gesture Generation

Co-speech gestures relate to contexts of speech content, audio, interacting persons, and so on. Among them, speech content is the primary context in studies of co-speech gesture generation owing to its relevance and importance [2]. A straightforward method for making gestures from speech content is to associate a speech word to a gesture [8]. A NAO robot adopted this strategy, so it made predefined gestures for selected words [9]. For example, the robot showed a deictic gesture when it said "you" and a metaphoric gesture for the word "every." The rule-based method is easy to implement, yet significant human labor is required to build a set of association rules. To overcome the drawbacks of the rule-based method, M. Kipp proposed to learn gestures from human demonstrations [10]. He built a probabilistic model for gesture generation, and the model was trained on a dataset of human gestures. The learning-based method resembled how humans, especially babies, learn social behaviors [11]. Another study proposed an improved graphical model that was demonstrated on a robot [12]. However, owing to the complexity of natural language and human body motions, the previous studies simplified the problem by training and generating only a small set of predefined gesture types. As a result, the generated gestures were simple and repetitive, and transitions between gestures were not natural. In the present paper, we represent gestures as a sequence of poses without the help of human experts' knowledge or additional annotation. The proposed model has a higher complexity, but it has much freedom at the same time.

Speech audio is a secondary context for gesture generation. Audio-driven gestures have been studied for artificial avatars in video games and virtual reality [13], [14]. In these applications, expressive speech audio, high-quality recordings of human voices, are available. Expressive speech audio, however, is usually unavailable from conventional TTS used in personal robots. Thus, in the present study, we do not use audio for co-speech gesture generation. Nonetheless, in implementing a robot prototype, synchronization between synthesized speech audio and generated co-speech gestures was investigated for the completeness of the study.

### B. End-to-End Learning for Motion Generation

Thanks to recent advancements in deep neural networks and abundant data, complex problems are being solved in an end-to-end manner without dividing the problem into smaller subproblems. Recently, motion generation studies started using end-to-end learning. Facial expressions while speaking were generated from raw speech audio [14]. Facial expressions were represented as thousands of 3D vertices, and the movements of the vertices were estimated frame by frame. Human motions of playing musical instruments were also generated from music [15]. In addition, human motions were generated from text describing a motion with a recurrent neural network [16]; this study was similar to ours in its form of mapping from text in natural language to human motions. However, in our problem, connections between texts and motions are much weaker and ambiguous, so they are difficult to learn. To the best of our knowledge, this is the first paper about end-to-end learning for generating co-speech gestures from speech text.

### III. TED GESTURE DATASET

Co-speech gestures are everywhere. People make gestures when they chat with others, give a public speech, talk on a phone, and even think aloud. Despite this ubiquity, there are not many datasets available. The main reason is that it is expensive to recruit actors/actresses and track precise body motions. There are a few datasets available (e.g., MSP-

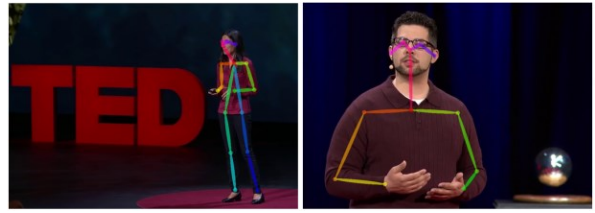

Figure 2. Samples of the TED Gesture Dataset. Extracted human poses are overlaid on the images.

AVATAR [17] and Personality Dyads Corpus [18]), but their sizes are limited to less than 3 h, and they lack diversity in speech content and speakers. The gestures also could be unnatural owing to inconvenient body tracking suits and acting in a lab environment.

Thus, we collected a new dataset of co-speech gestures: the *TED Gesture Dataset*. TED is a conference where people share their ideas from a stage, and recordings of these talks are available online. Using TED talks has the following advantages compared to the existing datasets:

- Large enough to learn the mapping from speech to gestures. The number of videos continues to grow.
- Various speech content and speakers. There are thousands of unique speakers, and they talk about their own ideas and stories.
- The speeches are well prepared, so we expect that the speakers use proper hand gestures.
- Favorable for automation of data collection and annotation. All talks come with transcripts, and flat background and steady shots make extracting human poses with computer vision technology easier.

### A. Collection and Annotation

The videos of TED talks were obtained from the official TED channel on YouTube. Outdated videos of low resolution and videos of music performances or interviews were excluded. In total, 1,295 videos and English transcripts were collected; the transcripts have timestamps for phrases. We extracted human poses for all frames by using *OpenPose* [19]. Then, each video was segmented into smaller shots. The TED talks consist of various shots of long, full, aerials, low angles, close-ups, etc., but what we are interested in is medium and medium-long shots showing upper-body gestures clearly. Shots of interest were selected under the following conditions: 1) the head, shoulder, arms, and hands of a speaker are visible; 2) the height of the upper body is larger than half of the video's height; 3) the speaker is facing near front; 4) shots are longer than 5 s; and 5) no still pictures. Fig. 2 shows examples of the selected shots.

The entire process of data collection and annotation is automated. To select shots of interest, the videos were segmented into shots by detecting sudden changes of motions or colors of images [20]. The segmented shots were filtered according to the rules described above. In the automated procedures, some errors in estimating human poses and shot segmentations were inevitable. To minimize the errors, we further removed the shots having missing joints or jittering poses.

## B. Statistics of the Dataset

TABLE I. STATISTICS OF TED GESTURE DATASET

| | |
|---|---|
| Number of videos | 1,295 |
| Average length of videos | 13 min |
| Shots of interest | 14,221 (11 per video on average) |
| Ratio of shots of interest | 12.9% (14,221 / 109,946) |
| Total length of shots of interest | 52.7 h |

The ratio of shots of interest is quite low since we selected shots conservatively to avoid bad samples that can mislead the learning process. The dataset is available publicly (TBA).

## IV. PROPOSED CO-SPEECH GESTURE GENERATION

### A. Data Representation and Preprocessing

In the present study, a speech text is represented as a sequence of words, and each word is encoded as a one-hot vector that indicates the word index in a dictionary. One-hot vectors are high-dimensional and sparse, so it is typical to convert them to compact representations, known as word embedding. In the space of word embedding, words of similar meaning have similar representations, so understanding natural language is easier. We used the pretrained word embedding model *GloVe*, trained on the Common Crawl corpus [21]. The dimension of word embedding is 300, and a zero vector is used for unknown words.

A gesture is represented as a sequence of human poses. We consider only the upper body, so each human pose is represented as a set of eight positions of the head (specifically the nose), neck, shoulders, elbows, and wrists as defined in *OpenPose* [19]. The poses were normalized so that the neck is at the origin and the length of the shoulder is 1. Similar to word embedding, human poses were converted to 10-dimensional vectors by using Principal Component Analysis (PCA), and 10 principal components explained 94.8% of the variance in the training dataset. As shown in Fig. 3, we can find meaningful human motions in the learned PCA space. For example, the second component is correlated to lowering both arms, and the last component shows open-arm gestures. In addition, we found that first and fourth principal components are related to in-plane rotations. Restraining rotations, not relevant to gestures, help with the learning process, so the values of the first and fourth dimensions were clipped to (-1, 1).

### B. Network Architecture

Co-speech gesture generation is a problem of mapping a sequence of words to a sequence of human poses. The problem resembles a neural machine translation in the form of sequence-to-sequence mapping. Inspired by neural machine translation research proposed in the *Seq2Seq* model [22], we propose a neural network consisting of an encoder and decoder, as shown in Fig. 4. The encoder processes input speech; it takes words one by one. A bidirectional recurrent neural network captures speech context, and the results are transmitted to the decoder to generate gesture motions. For decoding, we used a recurrent neural network with pre- and post-linear layers. A soft attention mechanism [23] was also used, so the decoder focused on specific words instead of whole text when it generated poses. Two-layered GRUs [24] with 200 hidden units were used for the encoder and decoder.

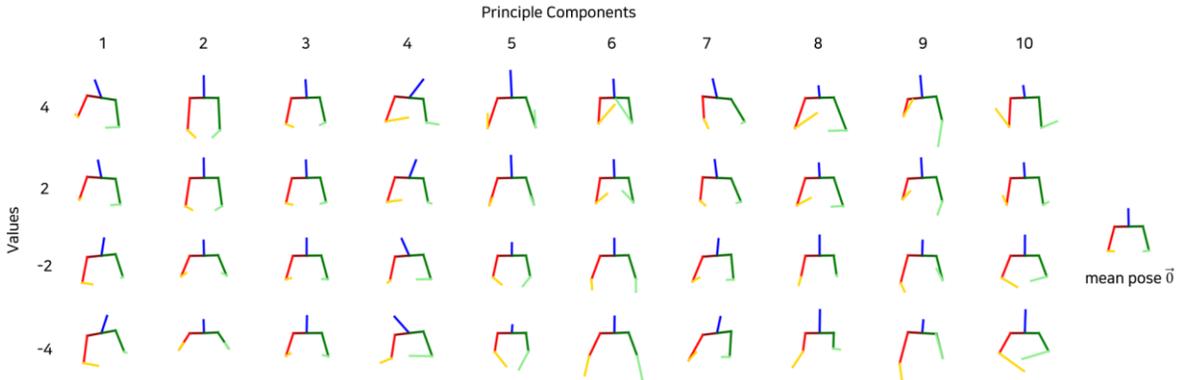

Figure 3. PCA subspace for upper-body poses. This figure shows how the human poses change according to the principal components. Different principal components and values are shown horizontally and vertically. The mean pose (i.e., zero vector) is also shown on the right side.

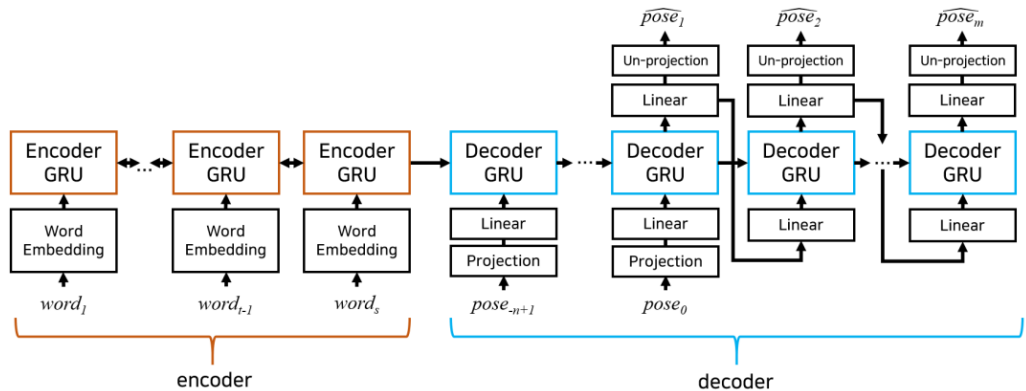

Figure 4. Proposed network architecture. The encoder GRU interprets $s$ speech words, and the decoder GRU generates $m$ human poses of gestures. The decoder GRU inputs $n$ previous poses to make the series of poses continuous. The soft attention mechanism is used but not depicted here.

Training a recurrent neural network on long sequences of more than hundreds of steps is usually not feasible owing to gradient vanishing or exploding. Therefore, we designed the network to generate a limited number of motion steps. For a long speech text, the network was inferred multiple times and the resulting sequences were concatenated. The decoder inputs *n* previous poses and generates *m* successive poses; this configuration makes the output poses of multiple inferences smooth. The parameters *n* and *m* were fixed to 10 and 20, respectively, in all experiments.

*C. Training*

We defined the loss function as

$$\mathcal{L} = \mathcal{L}_{mse} + \alpha \cdot \mathcal{L}_{continuity} + \beta \cdot \mathcal{L}_{variance}. \quad (1)$$

$\mathcal{L}_{mse}$ is a mean squared error between the output poses and ground truth poses in the training dataset. $\mathcal{L}_{continuity}$ is introduced for continuity in successive poses:

$$\mathcal{L}_{continuity} = \frac{\sum_{t=2}^{m}\|p_t - p_{t-1}\|}{m-1}, \quad (2)$$

where $p_t$ is a pose at time *t*. $\mathcal{L}_{variance}$ is defined as the negative of the variance of $p_t$, so it guides the network to generate dynamic motions. Two parameters $\alpha$ and $\beta$ controls the weights of the loss terms, and they were empirically determined and fixed to 0.01 and 1, respectively. With these values, the three loss terms have similar orders of magnitude.

For training, 34,469 pairs of sequences of words and poses, sampled from the training set of the TED dataset, were used. We used Adam optimization with a learning rate of 0.0001. The batch size was 64, and gradients were clipped to (-5, 5) to prevent gradient exploding. A dropout rate of 0.1 was applied to the first layers in GRUs. The network was trained for 560 epochs until the loss did not decrease, and the training took about 22 h with a NVIDIA GTX 1080.

*D. Results*

The trained network generated various gestures according to different speech texts not in the training set. We found iconic gestures depicting actions, metaphoric gestures for abstract concepts, and deictic and beat gestures. Fig. 5 shows examples. In the first example, the network generates the iconic gesture depicting "hold in your hand." The second example shows the metaphoric gesture of widening arms for a concept of "all." The third and fourth examples show how the trained network generates different motions for two similar sentences. The network successfully captured distinctive words and generated a metaphoric gesture for "big" and a deictic gesture for "you" and "me." We also investigated the attention of the decoder while it generates poses. As shown in Fig. 5 (b), the decoder sees the words in order. Speech and gestures are supposed to be synchronized in a timely manner, so we believe the attention map indicates that the network is successfully trained to have synchrony without explicit guidance.

V. EVALUATION

Evaluating a generative model is challenging. It is not adequate to measure value-level differences between the original and generated outputs since this cannot capture the quality of generation correctly. For example, when we describe an important concept, widening arms, raising a hand, and a metaphoric gesture of holding something with hands are all adequate, but these gestures have large position-level differences. Thus, similar to an image generation study [25], we conducted a subjective evaluation to measure anthropomorphism (i.e., the generated gestures are human-like), likeability (i.e., people like the generated gestures), and speech-gesture correlation (i.e., gestures match the speech content).

*A. Methods*

We compared the proposed method to the ground truth and three baseline methods of random, nearest neighbor, and manual. Here are the implementation details:

**Ground truth (GT)** -- uses gestures of human speakers in the TED dataset, but the playback speed of the gesture motions is adjusted to match the duration of the synthesized speech audio. Note that we considered the extracted human poses by *OpenPose* as the ground truth.

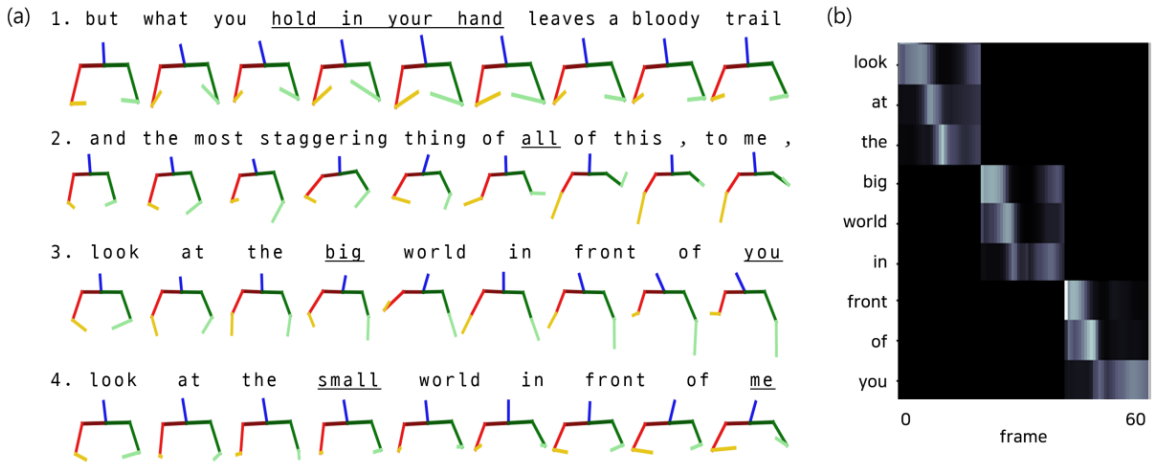

Figure 5. (a) Qualitative results. There are different gestures according to the speech context. For the speech of "hold in your hand," the stick figure makes the iconic gesture depicting holding hands. In the third sample, the metaphoric gesture for "big" and the deictic gesture for "you" are generated. The third and fourth samples demonstrate how the gestures differ for two similar sentences having different meanings. Distinctive words are underlined. (b) Attention map for the third sentence in (a). We tailored the attention maps for three inferences into one figure for better visibility.

**Nearest Neighbor (NN)** -- this method finds the most similar text in the training set, and the sequence of poses associated with the text is used. This baseline method was used in [26], and we modified it for our problem. BLEU [27] was used to measure text similarity. In this method, we divided an entire text into smaller chunks for better text matching, and concatenated the selected pose sequences.

**Random** -- selects a sequence of gesture poses from the training set at random. The sequence was tailored to have the same duration as the speech audio.

**Manual** -- the authors designed a sequence of gestures manually. We tried to make it fit to any speech context by using beat gestures and common metaphoric gestures (e.g., *Cup*, *Frame,* and *Emerge*) according to the gesture usage analysis in [10].

The baselines are competitive methods. The random and NN methods use human gestures in the training set, so the motions themselves should be smooth and human-like. For the NN, we interpolated poses of two consecutive chunks since there are motion discontinuities owing to dividing into chunks. In addition, in the evaluation, we used different gesture sequences of random and manual methods for different sentences to increase the variability and remove random effects. The gestures were demonstrated with stick figures to reduce appearance bias.

### B. Participants and Procedure

The participants were recruited from Amazon Mechanical Turk. To avoid gaming workers, we excluded participants who could not pass attention check questions or gave too-vague answers to questions about subjective impressions. Inconsistent answers, having opposed responses for two similar questions in an index group, were also rejected. We excluded 18 of 64 participants, so there were 46 valid participants. Half of them were female, and their ages were between 23 and 70 (M = 37). They were from the USA except for one from Australia, and they all were native English speakers or had bilingual proficiency.

In the evaluation, the participants viewed video clips demonstrating the proposed method, GT, and three baseline methods, and evaluated them subjectively. The videos had speech audio generated by using Google Text-to-Speech. There were no subtitles in the videos, but we asked the participants to read transcripts before the start of the evaluation. We used five sentences sampled from the test set of the TED dataset at random[1]. Each subject evaluated the methods for two sentences. The evaluation was done online, and took about 40 min. Although the presentation order of the methods was counterbalanced across the participants using Latin squares to prevent ordering effects, post-exclusion of participants breached the counterbalancing. However, the presentation orders were used a similar number of times (six to nine).

We used a questionnaire to measure three indexes of anthropomorphism, likeability, and speech-gesture correlation. The two first indexes are from the Godspeed questionnaires [28], and the last index was designed by the authors. In each index, there were three to five questions (Table II). All questions were answered using five Likert scales. The questions were shuffled randomly, and we flipped scales at random (e.g., Fake = 1, Natural = 5 or Natural = 1, Fake = 5).

TABLE II.  QUESTIONAIRE ITEMS USED TO EVALUATE GESTURES

| Index | Questionnaire Items |
|---|---|
| Anthropo-morphism | (Fake - Natural), (Machinelike - Humanlike), (Unconscious - Conscious), (Artificial - Lifelike), (Moving rigidly - Moving elegantly) |
| Likeability | (Dislike - Like), (Unfriendly - Friendly), (Unkind -Kind), (Unpleasant - Pleasant), (Awful - Nice) |
| Speech-Gesture Correlation | (Motions and speech are independent - Motions and speech are correlated), (Gestures ignore content - Gestures reflect content), (Gestures are not necessary - Gestures help to understand content) |

### C. Results

We first assessed that the indexes were reliable by measuring internal consistency. Cronbach's $\alpha$ of 0.93, 0.94, and 0.74 for anthropomorphism, likeability, and speech-gesture correlation, respectively, were all acceptable (> 0.6).

Fig. 6 summarizes the results of the evaluation. GT showed the best results for all indexes, and the proposed method showed the second-best results. The NN and manual methods showed similar results, but NN was rated slightly higher. Random was the worst of all indexes. According to ANOVA tests, there was a significant effect of methods on scores for the indexes of anthropomorphism and speech-gesture correlation [$F(4, 330) = 2.74$, $p = 0.03$, $F(4, 330) = 5.45$, $p < 0.01$], but not for the index of likeability [$F(4, 330) = 1.87$, $p = 0.11$]. Post hoc tests by using Fisher's Least Significant Difference indicated that the proposed method was rated significantly higher than the random for the anthropomorphism, likeability, and speech-gesture correlation indexes ($p = 0.009, 0.019, <0.001$). There were no statistically significant differences between the proposed and NN methods for the three indexes ($p = 0.09, 0.37, 0.20$) and between the proposed and manual methods ($p = 0.07, 0.33, 0.06$) with an error level of 0.05.

## VI. ROBOT PROTOTYPE

The previous sections demonstrated the co-speech gesture generation model with a stick figure, which is a simplified version of the human skeleton. However, we cannot use stick figures directly for a robot prototype that makes gestures while speaking. This section describes how we bring a stick figure to reality. The overall procedure is depicted in Fig. 7.

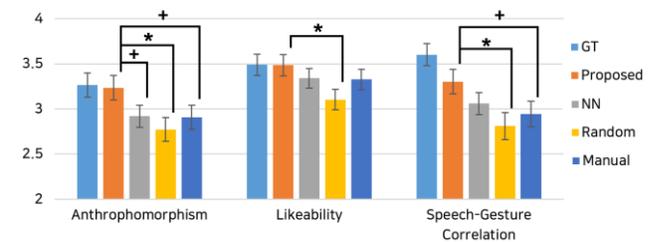

Figure 6.  Means and standard errors of methods for the evaluation indexes. Statistical differences between the proposed method and others are denoted with markers (+ $p < 0.1$; * $p < 0.05$). The graph is best viewed in color.

---

[1] Randomly selected clips for the evaluation: (1) youtu.be/Wai4ub90stQ (8:41-9:01), (2) youtu.be/27lMmdmy-b8 (1:00-1:14), (3) youtu.be/kcEIsbO0ivA (4:36-4:54), (4) youtu.be/CR_LBcZg_84 (0:58-1:45), (5) youtu.be/ZX8MBBohX3s (0:18-0:41)

The first step is 3D pose estimation from stick figures. There are existing studies for 3D pose estimation, but we found that they were not successful in the TED dataset owing to environment mismatches. Therefore, we implemented a neural network for our purpose. The task is converting 2D poses into 3D poses. This is easier than conventional 3D pose estimation since all humans are facing near front, and co-speech gestures are far less dynamic than sports motions. The network consists of a cascade of three fully connected layers with 30, 20, and 7 nodes with batch normalization. It estimates depth values for the input of upper-body joints in 2D. For a training dataset, we used the CMU Panoptic Dataset [29], which provides highly accurate 3D poses of social activities including many co-speech gestures. There were 604,190 frames in the training set, and the data were further augmented with rigid-body rotations and random noises on the joints.

Estimated 3D poses need to be retargeted to the robot. In the present study, we used the humanoid robot NAO from Softbank Robotics. The robot and 3D stick figures have the same joint configurations for the upper body, so we simply copied the joint angles for retargeting. The robot has 12 degrees of freedom: pitch and yaw of the head, pitch and roll of L/R shoulders, roll and yaw of L/R elbows, and yaw of L/R wrists. Joint angles were calculated analytically from 3D poses except for the pitch of head and yaw of wrists. We set the pitch of head and yaw of wrists to zero because these cannot be calculated without poses of face and hands.

The robot should make gestures while speaking a given speech text, so aligning gestures and speech audio is necessary. Algorithm 1 shows the overall procedure. First, speech is synthesized by using the Google TTS API. Then, the input text is split into several chunks, which contain a few words for a single inference of the trained network. The number of words in a chunk is determined by considering the frame rates of motions in the training dataset. Finally, sequences of poses are generated from the chunks of words. The generated motions have the same length of speech, and the robot plays gestures while speaking the synthesized speech.

The robot was able to generate gestures without observable differences from the 2D poses (see the supplementary video). In addition, we can easily apply the model to other robot platforms having human-like joint configurations since the intermediate representation of 3D poses is not dependent on the robot platform. The processing time was minimal. The network inferences were completed in 0.14 s in a CPU for seven inferences for a sentence of 25 words.

## VII. DISCUSSION AND LIMITATIONS

We proposed an end-to-end model for co-speech gesture generation from speech text. The model was trained on a TED dataset without prior knowledge about co-speech gestures, and showed successful results of showing several types of gestures appropriate to speeches. The trained model was also able to generate continuous gestures for any speech text of any length. The proposed method was better than the baselines in the subjective evaluation for all indexes of anthropomorphism, likeability, and speech-gesture correlation. We found that the indexes of anthropomorphism and speech-gesture correlation are important in co-speech gestures; the participants in the evaluation support this as follows: "positive impressions were human-like movements not stiff, moving freely"; "when the robot arms are not moving in a predictable human fashion, it actually hurts the experience"; "I had a positive impression when the speech correlated with the motions"; and "positive impression when I felt like the motions flowed smoothly with the content of what was being said."

The proposed method, with a loss term increasing motion variance, sometimes made excessive gestures. Several participants disliked the exaggerated motions. They commented as follows: "I got a negative impression if the gesture was too 'jerky' or fast," and "looked much more brash ... jumped around from motion to motion." One participant suggested that a few but clear gestures are better than incessant gesturing.

In this study, we considered only speech text and not audio. Therefore, the generated gestures and speech audio could not be tightly coupled. A few participants pointed out that "gestures were faster than speech, and this made it look unnatural." Our approach can be extended to generate gestures and speech audio together. This can generate prolonged phonemes when a robot makes long gestures for an important word, and gesture and speech audio should be tightly synchronized. Another direction of the extension is personalization. A variety of gestures according to speech context were demonstrated, but the gestures were the same for all robots. Parameters of controlling expressiveness, cultural dependency, and politeness would be beneficial for social robots.

## ACKNOWLEDGMENT

Thanks to Kyoungyong Park for preparing and managing the online evaluation.

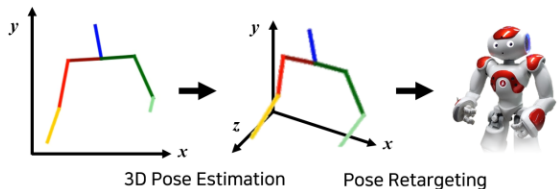

Figure 7. Gesture generation procedures for a robot prototype. Generated poses in 2D are transformed into 3D poses, and then the 3D poses are retargeted to a humanoid robot.

**Algorithm 1** Making Gestures While Speaking

$m$ = # of estimating poses, $n$ = # of previous poses,
$S$ = # of words of input text
$frame\_duration = 1/12$  // assumed 12 FPS
**Input:** speech sentence; $text = \{word_1, ..., word_S\}$
1: Synthesize speech audio; $speech = \text{TTS}(text)$,
   $speech\_duration = \text{get\_speech\_duration}(speech)$
2: Calculate a number of words for an inference chunk;
   $s = [S \times (m + n) \times frame\_duration / speech\_duration]$
3: Split text into inference chunks;
   $chunk_i = \{word_{i \times s + 1}, ..., word_{(i+1) \times s}\}$
4: **for each** $chunk_i$ **do**
5:   $pose_{i \times m+1, ..., (i+1) \times m} = gesture\_inference(chunk_i)$
6: **end for**
7: Play $pose$ and $speech$  // they have same duration